\documentclass{article}

\usepackage[utf8]{inputenc} % allow utf-8 input
\usepackage[T1]{fontenc}    % use 8-bit T1 fonts
\usepackage{hyperref}       % hyperlinks
\usepackage{url}            % simple URL typesetting
\usepackage{booktabs}       % professional-quality tables
\usepackage{amsfonts}       % blackboard math symbols
\usepackage{nicefrac}       % compact symbols for 1/2, etc.
\usepackage{microtype}      % microtypography
\usepackage{graphicx}

\title{Predictor-Corrector(PC) Temporal Difference(TD) Learning (PCTD)}

\author{Caleb Bowyer\\
Department of Electrical and Computer Engineering\\
University of Florida\\
  Gainesville, FL \\
  \texttt{c.bowyer@ufl.edu} \\
}
\begin{document}

\maketitle

\begin{abstract}
Using insight from numerical approximation of ODEs and the problem formulation and solution methodology of TD learning through a Galerkin relaxation, I propose a new class of TD learning algorithms. After applying the improved numerical methods, the parameter being approximated has a guaranteed order of magnitude reduction in the Taylor Series error of the solution to the ODE for the parameter $\theta(t)$ that is used in constructing the linearly parameterized value function. Predictor-Corrector Temporal Difference (PCTD) is what I call the translated discrete time Reinforcement Learning(RL) algorithm from the continuous time ODE using the theory of Stochastic Approximation(SA). Both causal and non-causal implementations of the algorithm are provided, and simulation results are listed for an infinite horizon task to compare the original TD(0) algorithm against both versions of PCTD(0).
\end{abstract}
\tableofcontents

\section{Introduction}
This novel family of reinforcement learning algorithms will stay within the realm of Eulerian approximations to the parameter ODE which is used to construct a TD-like learning algorithm, but will show to be a strict improvement over the explicit Eulerian based algorithms found in \cite{BVR} by an order of magnitude in the step size of the algorithm. 

Firstly, I will rederive "TD(0) learning" for the discounted cost objective. Discounted cost is the only objective considered in this paper although the ideas carry over to other cost settings. Additionally, a finite state space $X=\{1,2,...,n\}$ will be used for all developments. An uncontrolled irreducible aperiodic Markov chain samples points from this state space. To understand the TD(0) algorithm, one must first understand the discounted Poisson equation:
\begin{equation}\label{eq:1}
 h(x) = c(x)+\beta Ph(x) = c(x)+\beta  E[h(X_{k+1})|X_k=x] 
\end{equation}

where $h(x)$ is our relative cost value function and $c(x)$ is the cost incurred by being in state $x$. Lastly, $P$ is the transition matrix for the Markov chain that also functions as the conditional expectation on the function $h(x)$.

Rewriting, we have 
$$0=E[-h(X_k) + c(X_k) + \beta h(X_{k+1})|\mathcal{F}_k],$$ where $\mathcal{F}_k = \{\sigma{(X_i):i\leq k}\}$
is our sigma algebra or history up to time $k$. The above step follows from the Markov property.

TD learning is based on a linearly parameterized family of value functions $h^{\theta}: X\mapsto \mathbb{R}$ with a Galerkin relaxation applied to approximate the conditional expectation. For the linear parameterization, let 
\begin{equation}\label{eq:2}
h^{\theta}(x) = \theta^T\phi(x) = \sum_{i=1}^d \theta_i\phi_i(x),
\end{equation}

where $d$ is some integer, usually much smaller than the state space size and $\phi_i$  for $i=1,2,...,d$ are the basis functions. As in TD(0), I use $\zeta_k = \phi(X_k)\in \mathbb{R}^d$ for our eligibility vectors. Notation is now introduced for the Bellman error as $\epsilon(X_k,X_{k+1}) = -h(X_k) + c(X_k) + \beta h(X_{k+1})$. 
It is a known fact that a Galerkin relaxation can be applied to the above conditional expectation:
\begin{equation}\label{eq:3}
E[\epsilon(X_k,X_{k+1})\zeta_k]=0,
\end{equation}
for any bounded $\mathcal{F}_k$ measurable R.V. $\zeta_k$.

Now, using \ref{eq:2} and \ref{eq:3} in \ref{eq:1}, one gets:
% \begin{eqnarray*}
% E[-h^{\theta}(X_k)+c(X_k)+\beta h^{\theta}(X_{k+1})|\mathcal{F}_k] &=& \\ E[-\phi^T(X_k)\theta + c(X_k) + \beta\phi^T(X_{k+1})\theta|\mathcal{F}_k] &=& \\ \left E[[-\phi^T(X_k)\theta + \beta\phi^T(X_{k+1})\theta + c(X_k)]\zeta_k]\right|_{\theta=\theta^*} &=& 0,
% \end{eqnarray*}
\begin{eqnarray*}
E[-h^{\theta}(X_k)+c(X_k)+\beta h^{\theta}(X_{k+1})|\mathcal{F}_k] &=& \\ E[-\phi^T(X_k)\theta + c(X_k) + \beta\phi^T(X_{k+1})\theta|\mathcal{F}_k] &=& \\  E[[-\phi^T(X_k)\theta + \beta\phi^T(X_{k+1})\theta + c(X_k)]\zeta_k] |_{\theta=\theta^*} &=& 0,
\end{eqnarray*}

where the last equality is achieved at the optimal parameter $\theta^*$ which is the root of the equation.

From this, I define the parameterized Bellman error as $$\epsilon^{\theta}(X_k,X_{k+1}) = [(\beta\phi^T(X_{k+1})-\phi^T(X_k))\theta + c(X_k)].$$ This expression is also called the temporal difference sequence: $$d_{k+1} = (\beta\phi^T(X_{k+1})-\phi^T(X_k))\theta_k + c(X_k).$$ The TD(0) algorithm is often presented as:
\begin{equation}
\theta_{k+1} = \theta_k + \alpha_k \zeta_k d_{k+1} 
\end{equation}\label{eq:4}
without any of the steps to the derivation presented above.

Now, consider how the TD(0) algorithm can be derived from the analysis of a particular ODE and leads to a root finding problem with expectation being taken in steady-state. It is a two part process: 

Step 1: Consider an ODE we wish to solve
\begin{equation}\label{eq:5}
\frac{d\theta(t)}{dt} = \alpha\bar{f}(\theta(t)),
\end{equation}
$\alpha > 0$ where 
\begin{equation}
\bar{f}(\theta(t)) = E_{\infty}[\zeta_k(-h^{\theta}(X_k) + c(X_k) + \beta h^{\theta}(X_{k+1}))]
\end{equation}
What has been done for the last 30 years or so has been to take this fixed point equation and apply explicit or direct Euler method to the ODE.
\newline
\newline Step 2a) Translate to discrete time 
\begin{equation}
\theta_{k+1} = \theta_k + \alpha_k\bar{f}(\theta_k)
\end{equation}
Step 2b) The expectation is dropped when applying SA
\begin{equation}
\theta_{k+1} = \theta_k + \alpha_k [\zeta_k(-h^{\theta}(X_k) + c(X_k) + \beta h^{\theta}(X_{k+1}))]
\end{equation}

This has led to the well known family of TD learning algorithms and later development of the control algorithms: SARSA and Q-learning. For more details on the ODE design and connection to the classical discrete time algorithms see \cite{Meyn2000ode} and the connection with Q-learning as well as more function approximation considerations in the asymptotic case \cite{melo2008analysis}.  In this paper, I am proposing to make all of the above algorithms more efficient in learning the optimal value function faster by improving the updates to the parameter $\theta_k$ with a better translation step from the algorithm design. 

Furthermore, I will provide conditions upon which the algorithm is stable in the convergence of the parameter $\theta_k$ to $\theta^*$, that results in learning the optimal value function. In a separate research direction, some work on nonlinear parameterized value functions has been done, but it is often dismissed because of divergent counterexamples found in the literature. E.g., \cite{BVR} and more recently a study in \cite{achiam2019towards} investigate this type of function approximation. For linear value function learning in a non-asymptotic setting see \cite{Bhandari} and for the control learning algorithms of this flavor see \cite{Zou}.
Now, I will reformulate the basic problem of TD(0) from a modern viewpoint before deriving the PCTD(0) algorithms.

The goal of RL can be stated as trying to solve i.e., find the zeros of this equation: 
\begin{equation}
\bar{f}(\theta) = E_{\infty}[\zeta_k \epsilon^{\theta}(X_k,X_{k+1})]
\end{equation}
The goal is to find a $\theta^*$ s.t. $\bar{f}(\theta^*) = 0$. In other words, at the fixed point $\theta^*$ of the ODE:
% \begin{equation}
% \frac{d\theta(t)}{dt} = \left A\theta(t) + b \right|_{\theta(t)=\theta^*} = 0
% \end{equation}
\begin{equation}
\frac{d\theta(t)}{dt}  =  A\theta(t) + b |_{\theta(t)=\theta^*} = 0
\end{equation}

and so the deterministic recursion after applying SA produces a stable algorithm provided A is negative definite.

From TD(0) learning in the discounted case, one can obtain:
$$ A = -E[\phi(X_k)\phi(X_k)^T] + \beta E[\phi(X_k)\phi(X_{k+1})]$$ and $$b = -E[\phi(X_k)c(X_k)]$$ which forms our linear equation for $\theta^*$ of the form $\bar{f}(\theta^*) = 0 \Longrightarrow A\theta^*+b = 0$ implies the solution is easily obtained from solving the system of equations. Provided A is full rank implies a sufficient condition for learning the optimal parameter vector $\theta^*$. Alternatively, the first matrix $E[\phi(X_k)\phi(X_k)^T]$ must form a complete linear independent basis, and because of the contraction term with $\beta$, that ensures stability of TD(0). 

A note on step-size of the algorithm: 
The step-size sequence in the final version of the translated algorithms will take the form of $\alpha_k = \frac{g}{k+n_0}$ with $g,n_0>0$. This is sometimes done to tune the step size to ensure optimal asymptotic covariance of the parameter estimate. The gain sequence, $\alpha_k$ whether it is decayed or a fixed learning rate $\alpha$ must be tuned for the problem at hand. See \cite{Devraj} for a complete exposition. The form of the above varying step-size has been shown to handle transients and steady-state performance through appropriate choice of $n_0$ and $g$. 

\section{Numerical Method Motivation}
In these pages, I propose and derive a new family of RL algorithms that come from a better study of the numerical methods that lead from the translation of the continuous time ODE to the final discrete time RL algorithm.
Step 1: Create your ODE as in \ref{eq:5} which will be used for the rest of this paper, although the methods and algorithms clarified and elaborated in this paper can be adapted to more complicated ODEs or cost objectives for the RL solution of choice.
Step 2 is what will be improved from previous research. 
Denote 
\begin{equation}
f(\theta) = [\zeta_n (-h^{\theta}(X_n) + c(X_n) + \beta h^{\theta}(X_{n+1}))]
\end{equation}

From the numerical ODE literature it is known that the predictor-corrector method performs better than explicit Euler because it cancels out the leading order Taylor Series error (T.E.)
First let us review explicit Euler and Modified Euler Methods. The Taylor Series(T.S.) expansion of 
$$\dot{\theta}=f(\theta,t)$$ is 
$$\theta(t)=\theta_0 + \theta_0'(t-t_0) + \frac{1}{2!}\theta_{0}''(t-t_0)^2+...$$

For small $\alpha = (t_1-t_0)$ if we just keep the $\mathcal{O}(\alpha)$ term then we obtain
\begin{eqnarray*}
    \theta(t_0+\alpha) &=& \theta(t_0) + \alpha\theta_0'\\
\end{eqnarray*}
This implies that
\begin{eqnarray*}
\theta_{n+1} &=& \theta_n + \alpha\theta_n'+ \mathcal{O}(\alpha^2)
              = \theta_n + \alpha f(t_n,\theta_n) + \mathcal{O}(\alpha^2) \forall n = 0,1,2,...
\end{eqnarray*}             

Euler's method is explicit. It is easy to implement, but inaccurate. 

% Finite difference interpretation:\newline
% $$\textit{ Forward difference: }\theta_n' = \frac{\theta_{n+1}-\theta_n}{h} + \mathcal{O}(h)$$
% $$\textit{ ODE(Exact): }\theta_n'=f(t_n,\theta_n)$$ 
% Combine: 
% $$f(t_n,\theta_n) = \frac{\theta_{n+1}-\theta_n}{h} + \mathcal{O}(h)$$
% Result: 
% $$\theta_{n+1} = \theta_n + hf(t_n,\theta_n) + \mathcal{O}(h^2)$$
% Geometric View Point: 
% $\theta_{n+1} = \theta_n + h*\textbf{Slope at}   (t_n,\theta_n)$\newline
% with Run: $h$, Rise: $h\theta_n'$, and Error: $\frac{h^2}{2}\theta_n''$
We see that the direct Euler's method has only a first order accuracy.
The next development before I can set up the new algorithm is to review the Euler Implicit method:
Evaluate the slope $f$ at $(t_{n+1},\theta_{n+1})$ \newline
$$\frac{\theta_{n+1} -\theta_n}{\alpha} + \alpha\theta_{n+1}'' = f(t_{n+1},\theta_{n+1})$$
$$ \Longrightarrow \theta_{n+1} = \theta_n + \alpha f_{n+1} - \alpha^2 \theta_{n+1}''= \theta_n + \alpha f_{n+1} + \mathcal{O}(\alpha^2) $$
The local T.E. in the explicit method is $0.5\theta_n''\alpha^2$, and the local T.E. in the implicit method is $-0.5\theta_{n+1}''\alpha^2$. Note the two terms are opposite to each other.
Major problems with both Implicit and Explicit Euler methods:
1) In general $f(\theta_{n+1},t_{n+1})$ depends on $\theta_{n+1}$ which is an unknown at time $n$.
2) If $f(\theta,t)$ is a non-linear function of $\theta$, the procedure for obtaining $\theta_{n+1}$ can be complicated using the implicit method. In both methods, the accuracy is poor. The T.E. is $\mathcal{O}(\alpha)$. This is where the modified Euler predictor-corrector method improves the accuracy of $\theta_n$. 

% The following is a deterministic example that can be used or modified to show power of predictor corrector over explicit Euler:
% \subsection{Toy Problem}
% $\theta' = -10(t + \theta)$. I could throw in a random variable if Dr. Meyn thinks that sort of example may be interesting. to compare convergence approaches from different ODES and compare different RL algorithms, or even have an RL algorithm track the solution? Maybe.

\section{PCTD(0) Derivation}
\subsection{Causal PCTD(0)}
PCTD(0) uses the improved numerical methods to improve the approximation to the optimal value function. In my notation for the derivation of PCTD(0) I will reserve the symbol h for the relative cost value function. For step-size, the symbol $\alpha$ will be used. One can combine the Euler Explicit and Implicit methods to cancel out the leading order T.E. 

Consider slope:
\begin{eqnarray*}
\frac{\theta_{n+1} - \theta_n}{\alpha} &=& \frac{1}{2}[f(t_n,\theta_n)+f(t_{n+1},\theta_{n+1})]\\
                                 &\equiv& \textbf{avg. slope over } [t_n, t_{n+1}]
\end{eqnarray*}
                                 
$$ \Longrightarrow \theta_{n+1} = \theta_n +\frac{1}{2}\alpha(f_n + f_{n+1}) + \mathcal{O}(\alpha^3)$$
In this way, the local T.E. is reduced to $\mathcal{O}(\alpha^3)$.
To obtain $f_{n+1}$ with $\theta_{n+1}$ yet to be determined, estimate $\theta_{n+1}$ for the next time period using the Explicit Euler at every step.
Predictor: 
\begin{equation}\label{eq:12}
\bar{\theta}_{n+1} = \theta_n + \alpha f_n + \mathcal{O}(\alpha^2)
\end{equation}
(As a preliminary predicted value of $\theta$ at $t_{n+1}$)\newline
Corrector: 
\begin{equation}\label{eq:13}
\theta_{n+1} = \theta_n + \frac{\alpha}{2}[f(t_n,\theta_n) + f(t_{n+1}, \bar{\theta}_{n+1})]
\end{equation}

Qualitative Error Analysis: When calculating $f_{n+1}$ in the corrector step, $\theta_{n+1}$ is replaced by $\bar{\theta}_{n+1}$
The modified Euler (Predictor-Corrector) method has $\mathcal{O}(\alpha^2)$ error. 

 To denote the linear value function's parameterization by $\theta$, it will be written as $h^{\theta}$. Note: 
 \begin{equation}
h^{\theta}(x) = \phi^T(x)\theta
 \end{equation}

After substituting, 
% \begin{equation}\label{eq:15}
%     f(\theta_n) = \left {(\zeta_n(-h^{\theta_n}(X_n)+c(X_n)+\beta h^{\theta_n}(X_{n+1})))}\right|_{\theta = \theta_n}
% \end{equation}
\begin{equation}\label{eq:15}
    f(\theta_n) =  [(\zeta_n(-h^{\theta_n}(X_n)+c(X_n)+\beta h^{\theta_n}(X_{n+1})))]|_{\theta = \theta_n}
\end{equation}

and
\begin{equation}\label{eq:16}
f(\bar{\theta}_{n+1}) = [\zeta_n (-h^{\bar{\theta}_{n+1}}(X_n) + c(X_n) + \beta h^{\bar{\theta}_{n+1}}(X_{n+1}))].
\end{equation}

The new improved TD learning algorithm called Causal PCTD(0) causal is:
\begin{equation}
\theta_{n+1} = \theta_n + \alpha_n \frac{1}{2}[f(t_n,\theta_n)+f(t_{n+1},\bar{\theta}_{n+1})].
\end{equation}

Futhermore the predictor is the same as in \ref{eq:12}.

% Then, the update looks like:
% \begin{eqnarray*}
% \theta_{n+1} = \theta_n + \frac{\alpha}{2}[\zeta_n[-\phi^T(X_n)\theta_n + c(X_n) +\beta\phi^T(X_{n+1})\theta_n]
%                             + \zeta_n[-\phi^T(X_n)\bar{\theta}_{n+1} + c(X_n) +\beta\phi^T(X_{n+1})\bar{\theta}_{n+1}]

% \end{eqnarray*}

From the definitions we can arrive at after substitution and reduction to the form
\begin{equation}\label{eq:18}
\theta_{n+1} = \theta_n + A\theta_n + b,
\end{equation}

where

$$A = \alpha\zeta_n[\beta\phi^T(X_{n+1})-\phi^T(X_n)] + \frac{\alpha^2}{2}[\zeta_n[\beta\phi^T(X_{n+1}) - \phi^T(X_n)]]^2$$

and 
$$ b = \alpha c(X_n)\zeta_n + \frac{\alpha^2}{2}\zeta_n(\beta\phi^T(X_{n+1})-\phi^T(X_n))\zeta_n c(X_n) .$$

After expanding, we have six terms remaining to analyze in the matrix $A$. Splitting $A$ into two matrices, it is easy to see what terms are involved: $A$ = $A_1$ + $A_2.$

$$A_1 = \alpha\beta\phi(X_n)\phi^T(X_{n+1}) - \alpha\phi(X_n)\phi^T(X_n) $$
and 
$$A_2 = \frac{\alpha^2\beta^2}{2} H^2 - \frac{\alpha^2\beta}{2}GH - \frac{\alpha^2\beta}{2}HG + \frac{\alpha^2}{2}G^2$$
where 

$G = \phi(X_n)\phi^T(X_{n+1})$ and $H = \phi(X_n)\phi^T(X_n) .$

$A_1$'s contribution to the dynamics are exactly what TD($0$) learning contributes to the learning algorithm. The remaining four terms come from the application of the Predictor-Corrector method to the ODE and are the additional terms contributed by Causal PCTD(0).

\subsection{Non-causal PCTD(0)}
Note that the time support for $\theta$ and $X$ are not exactly the same nor do they have to share the same time support entirely, only be relatively close. For example, at time $n+1$, $\theta_{n+1}$ will depend on both $(X_n,X_{n+1})$ the previous state and the present state. However, a minimal non-causal filter may improve the updates to $\theta_{n+1}$ drastically by using information from the past, present and future in the update of the parameter. I say the update is minimally non-causal in the sense that it only depends on the state information one time-step into the future. 

Now, I present the Non-causal PCTD(0) derivation taking this extra information into account so we can see the final update form:

Starting from \ref{eq:13} with definitions \ref{eq:12} and \ref{eq:15} the same, we redefine \ref{eq:16} to be:

\begin{equation}\label{eq:19}
f(\bar{\theta}_{n+1}) = [\zeta_{n+1} (-h^{\bar{\theta}_{n+1}}(X_{n+1}) + c(X_{n+1}) + \beta h^{\bar{\theta}_{n+1}}(X_{n+2}))].
\end{equation}

From these definitions, it is natural to see that a better update may result from using the triple: $(X_n, X_{n+1}, X_{n+2})$ in this form of the update, rather than using only past and present information. Practically, to apply this algorithm in real-time, you would only have to wait one time-step to use the updated improved parameter.

% Then,
% \begin{small}
% \begin{eqnarray*}
% \theta_{n+1} = \theta_n + \frac{\alpha}{2}[\zeta_n[-\phi^T(X_n)\theta_n + c(X_n) +\beta\phi^T(X_{n+1})\theta_n]
%                             + \zeta_{n+1}[-\phi^T(X_{n+1})\bar{\theta}_{n+1} + c(X_{n+1}) +\beta\phi^T(X_{n+2})\bar{\theta}_{n+1}]

% \end{eqnarray*}
% \end{small}

After substituting and reducing, we can again arrive at the same standard update form as in \ref{eq:18},
% $$\theta_{n+1} = \theta_n + A\theta_n + b$$

where $A = A_1 + A_2.$
For Non-causal PCTD(0), I split $A$ into two parts. $A_1$ will hold the causal terms, and $A_2$ will hold the non-causal terms.

$$A_1 = \frac{\alpha}{2}[-\zeta_n\phi^T_n - \zeta_{n+1}\phi^T_{n+1}+\beta\zeta_n\phi^T_{n+1}+ \alpha\zeta_{n+1}\phi^T_{n+1}\zeta_n\phi^T_n - \beta\alpha\zeta_n\phi^T_{n+1}\zeta_n\phi^T_{n+1} ]$$
and
$$A_2 = \frac{\alpha}{2}[ \beta\zeta_{n+1}\phi^T_{n+2} - \alpha\beta\zeta_{n+1}\phi^T_{n+2}\zeta_n\phi^T_n  + \alpha\beta^2\zeta_{n+1}\phi^T_{n+2}\zeta_n\phi^T_{n+1}]$$
and
$$ b = \frac{\alpha}{2}[\zeta_n c_n + \zeta_{n+1} c_{n+1}] - \frac{\alpha^2}{2}\zeta_{n+1}\phi^T_{n+1}\zeta_n c_n + \frac{\alpha^2\beta}{2}\zeta_{n+1}\phi^T_{n+2}\zeta_n c_n.$$

 Comments: this leads to a direct eigenvalue test for stability. Using the law of large numbers, the exact parameter $\theta^*$ can be computed only assuming a full rank condition on the basis functions. 
 
 \subsection{Extension to PCTD($\lambda$)}
 Futhermore, we can generalize these results to a generalized TD($\lambda$)-like algorithm, which I call PCTD($\lambda$). To obtain the PCTD($\lambda$) general versions of the new RL algorithms, one needs to substitute for the eligibility vector $\zeta_n = \phi_n + \beta \lambda \zeta_{n-1}$. This generalizes the algorithms further for any $\lambda \in [0,1)$, and stability is guaranteed under the same mild set of assumptions.

% An example of a floating figure using the graphicx package.
% Note that \label must occur AFTER (or within) \caption.
% For figures, \caption should occur after the \includegraphics.
% Note that IEEEtran v1.7 and later has special internal code that
% is designed to preserve the operation of \label within \caption
% even when the captionsoff option is in effect. However, because
% of issues like this, it may be the safest practice to put all your
% \label just after \caption rather than within \caption{}.
%
% Reminder: the "draftcls" or "draftclsnofoot", not "draft", class
% option should be used if it is desired that the figures are to be
% displayed while in draft mode.
%
% \begin{figure}[!h]
% \centering
% \includegraphics[width=0.8\columnwidth]{trilece} 
% \caption{Simulation Results}
% \label{fig_sim}
% \end{figure}

% Note that IEEE typically puts floats only at the top, even when this
% results in a large percentage of a column being occupied by floats.

\section{Prediction Task}
\begin{figure}[hbt!]
  \centering
{\includegraphics[height=6cm, width=10cm]{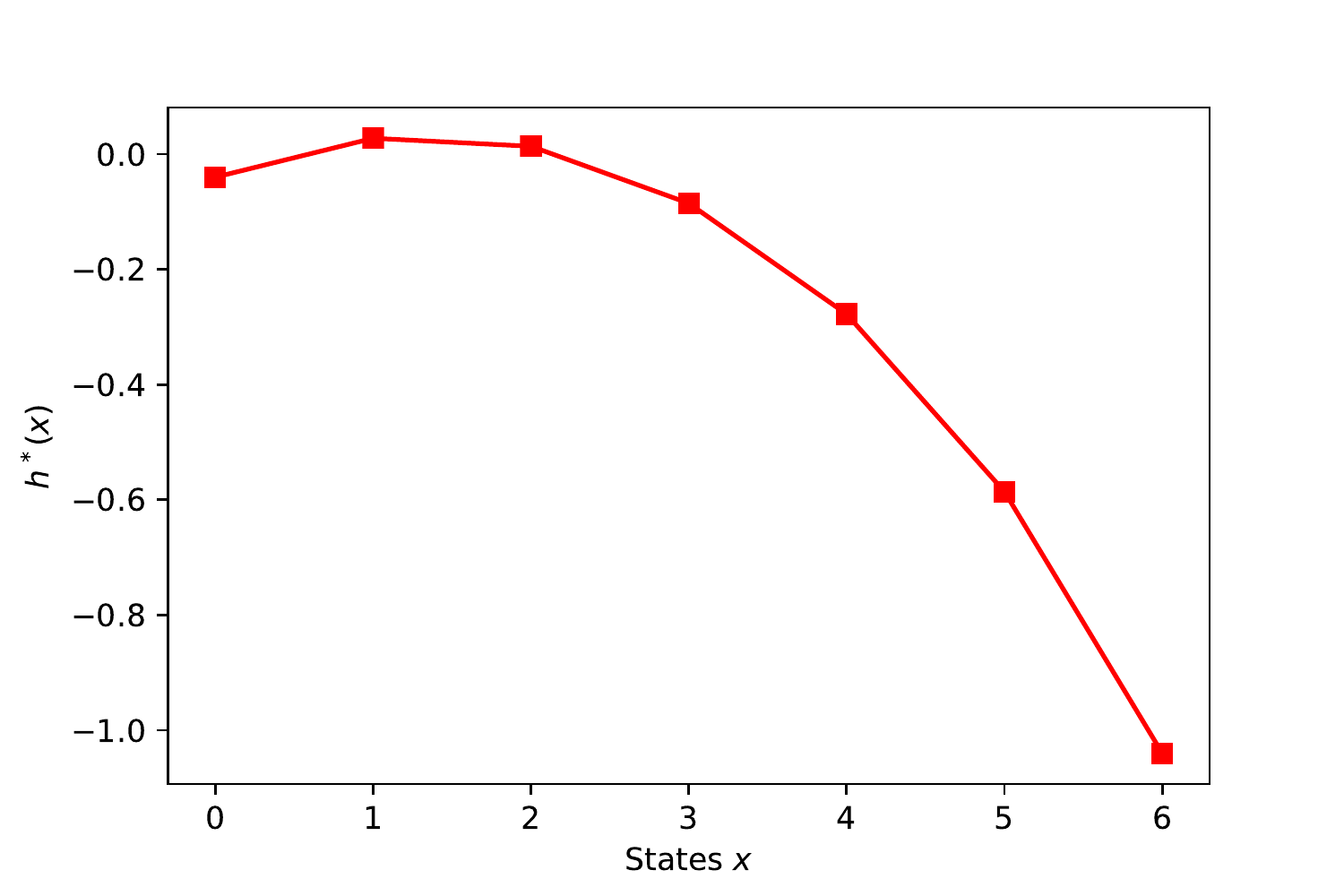}}
    \caption{Optimal Value Function $h^*(x)$ from sparse costs in the Prediction Task}
  \label{fig:sparse_rewards}
\end{figure}
For testing PCTD(0) varieties vs TD(0) learning I will be implementing an infinite state MDP random walk on 7 states. Two states are "terminal": the leftmost and rightmost states, i.e., once the agent reaches either x=0 or x=6 the agent directly returns to state x=3 in the next step with probability one. All independent games start in state x=3, and with probability $\frac{1}{2}$ the agent transitions left or right.  The learning environment is inspired by the random walk example from the introductory RL book \cite{Sutton} Read Ch.6 on Temporal Difference Learning. However, there it is formulated as episodic, which means the game ends once the agent reaches either x=0, or x=6. Then, in the next episode, the agent starts at x=3 whereas in my MDP, the play is unending. In the simulation section, the transient performance of the algorithms will be compared. Additionally, comparisons will be made between using a decayed learning rate and a fixed learning rate for both Causal and Non-causal PCTD(0) algorithms. The true optimal value function in \ref{fig:sparse_rewards} was obtained by running value iteration until the Bellman error reached zero with a $\beta=0.95$ discount factor.

% To reference figure use \ref{fig:sparse_rewards}

% \begin{figure}
%   \centering
% {\includegraphics[width=10cm]{Optimal_Value_Function_linear.pdf}}
%     \caption{Optimal Value Function  $h^*(x)$ from linear costs}
%   \label{fig:sparse_rewards}
% \end{figure}
\section{Simulation Results}
\begin{figure}[hbt!]
  \centering
{\includegraphics[height=6cm, width=10cm]{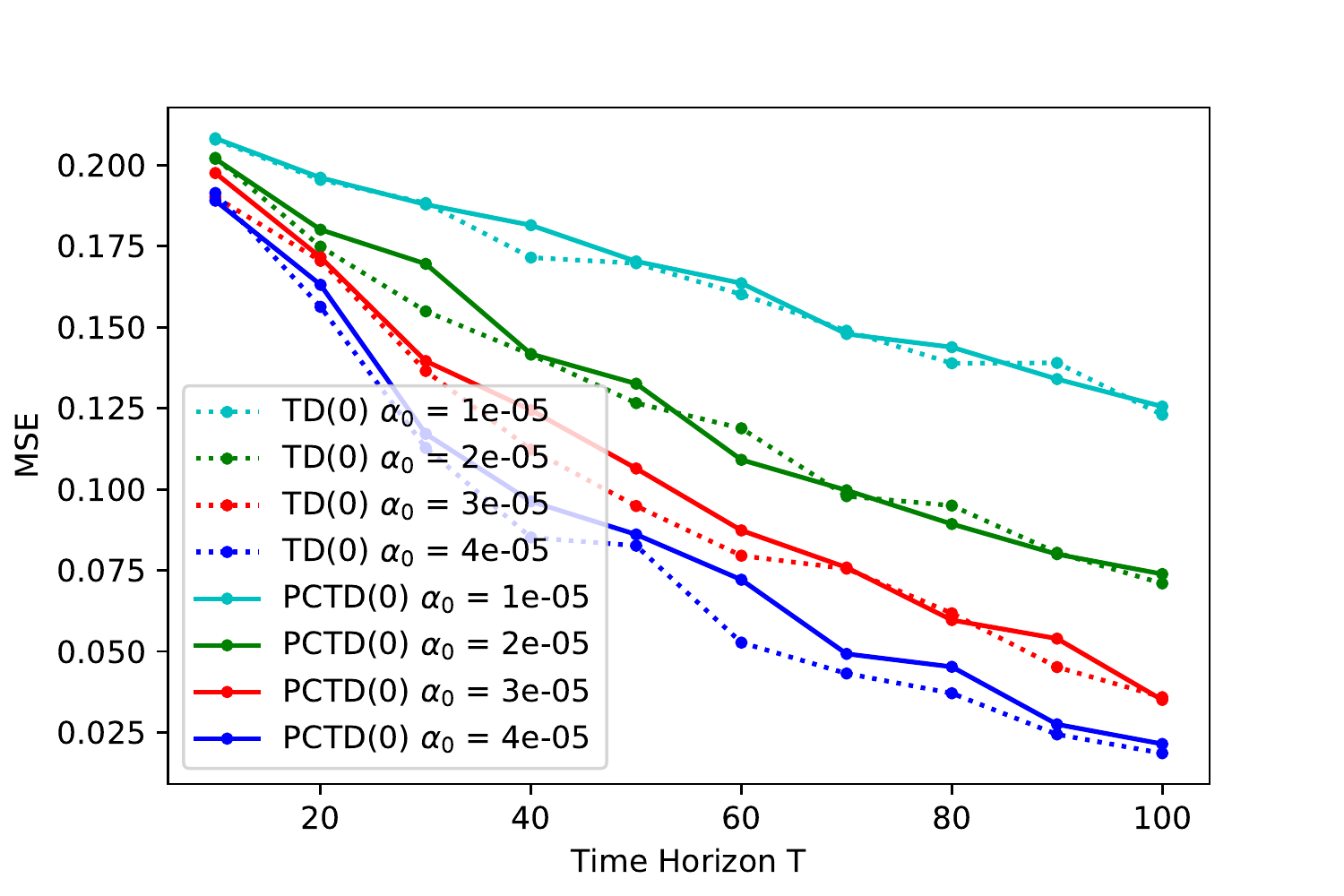}}
    \caption{Causal PCTD(0) for N=100 independent trials for decayed learning rates}
  \label{fig:causal}
\end{figure}

\begin{figure}[hbt!]
  \centering
{\includegraphics[height=6cm, width=10cm]{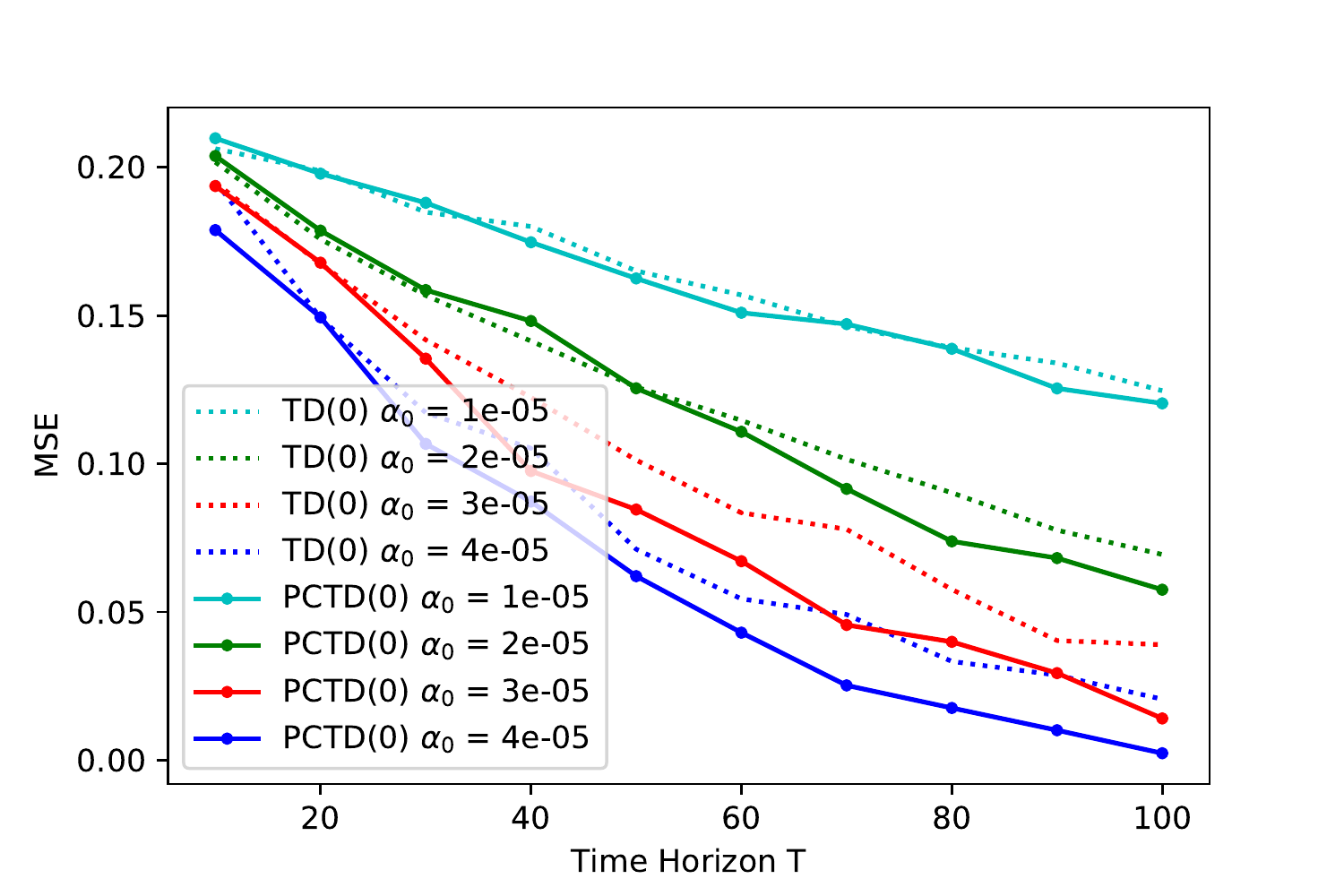}}
    \caption{Non-Causal PCTD(0) for N=100 independent trials for decayed learning rates}
  \label{fig:Noncausal}
\end{figure}
In this section we will test the performance in practice for the two versions of PCTD(0), both Causal and Non-causal to determine which leads to lower MSE during early learning for the value-function prediction task. 
For TD(0) and both versions of PCTD(0) a band of learning rates with decayed learning rates was found which resulted in stable behavior of the iterations. Below an initial learning rate of $1e^{-5}$ for the decayed learning rate sequences, stability was not guaranteed for any of the algorithms using our fixed polynomial basis of linear, quadratic, and cubic features of the form $[x, \frac{1}{2}x^2, \frac{1}{3}x^3]$. Hence $d=3$ was the number of features.

 Within the range of $\alpha_0 =  1e^{-5} - 1e^{-6}$ performance improved monotonically with decreasing initial learning rates.  Graphed in both figures \ref{fig:causal} and \ref{fig:Noncausal} is the MSE averaged over all states for the linearly parameterized value function averaged over 100 independent trials each of time extent $T = 100$ steps for both versions of PCTD(0) against TD(0). The performance of TD(0) vs. Causal PCTD(0) and TD(0) vs. Non-causal PCTD(0) on the sparse cost Prediction Task is illustrated. The causal version of PCTD(0) performed just as well as TD(0) on the prediction task whereas the Non-causal version of PCTD(0) offered a noticeable advantage at early learning of the optimal value function. These experiments were repeated using fixed learning rates, and essentially the same results were obtained. Lastly, I will show how the PCTD(0) and TD(0) value functions' estimates compare to the optimal value function by showing confidence bounds using $N=100$ independent experiments with $g = 0.01$, and $n0 = 200$ for the same Horizon $T=100$ to show clearly the early learning advantages of Non-causal PCTD(0) over TD(0) version at learning the optimal-value function $h^*(x)$.

\begin{figure}[hbt!]
  \centering
{\includegraphics[height=6cm, width=10cm]{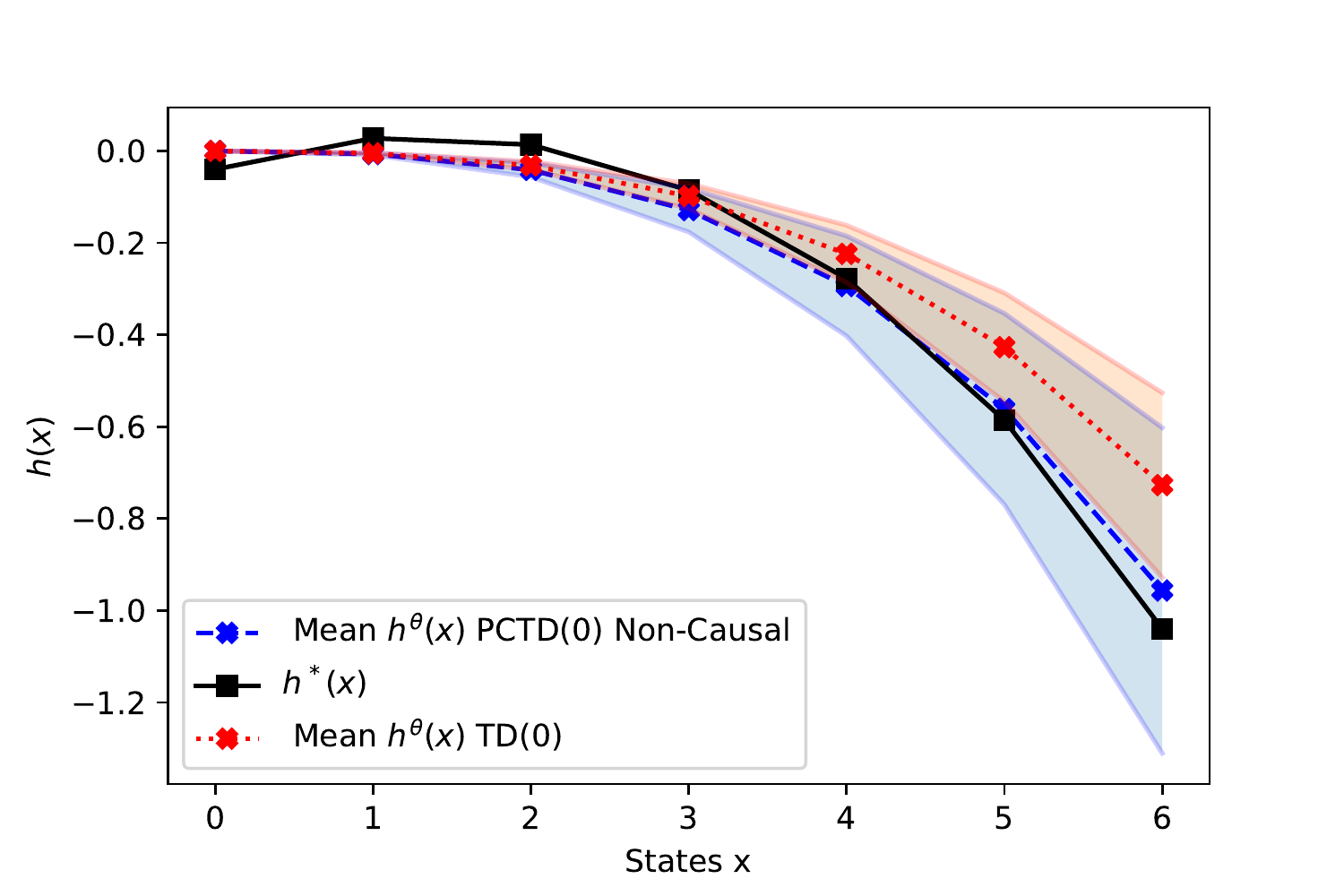}}
    \caption{$h(x)$ over $N=100$ independent runs for $T=100$}
  \label{fig:Noncausalvalues}
\end{figure}

% \begin{figure}
%   \centering
% {\includegraphics[height=4cm, width=10cm]{Causal_value_function.pdf}}
%     \caption{$h(x)$ over $N=100$ independent runs for $T=100$}
%   \label{fig:Noncausal}
% \end{figure}

\section{Conclusion and Future Work}
The Causal PCTD(0) performs slightly worse than TD(0) across the band of stable learning rates. Meanwhile, the Non-causal PCTD(0) variant appears to have faster transient learning capabilities for learning the optimal value function during the early stages of learning as compared to the TD(0) algorithm in terms of utilizing the past, present, and future information in its updates. It should be noted that if the stability of the algorithms could be made to improve with larger initial learning rates, the Non-causal PCTD(0)'s MSE would most likely decrease even faster than standard TD(0)'s MSE based on being more accurate in the parameter estimate by a factor of $\mathcal{O}(\alpha)$. To improve Non-causal PCTD(0)'s faster learning gains more, different types of feature vector's may show other advantages for this algorithm over standard TD(0) that increases its effectiveness than those features employed in the simulation section of this paper. 

Furthermore, the improvement in MSE of Non-causal PCTD(0) over Causal PCTD(0) suggests that Non-causal PCTD(0)'s derivation leads to the correct form of the update based on the Predictor-Corrector numerical ODE method being applied to the original ode \ref{eq:5}. This implies Non-causal PCTD(0) has the correct interpretation of the time support for $\theta$ and $X$ involved in the expression on the right hand side of \ref{eq:19}. Non-causal PCTD(0)'s form of algorithm seems to be able to admit further extensions that will be explored in future work such as incorporating additional future and past feature terms for every update of the parameter $\theta_{k+1}$. Additional numerical methods of higher order will also lead to further RL algorithm developments. In a future paper, I will seek to tie together different temporal difference learning algorithm around a common thread of improving numerical methods, such as the Runge-Kutta methods or even Richardson extrapolation. This will create a new family of temporal difference learning algorithms. Lastly, the PCTD algorithms developed in this paper will be studied in a deterministic setting later where the noise comes from deterministic probing signals, such as well structured mixtures of sine waves. In this future paper, the state space will be continuous, and the control space will be as well. The proof of convergence developed in this future work may shed light on the probabilistic setting and lead to alternative proofs of stability and convergence rates. 

\section{Broader Impact}
Because this research is largely theoretical in nature, no material applications are foreseen. Hence, this section is not applicable for the current work. 

%\section{References}
% \begin{thebibliography}{8}
\bibliographystyle{unsrt}
\bibliography{main}

\end{document}